# Stone Needle: A General Multimodal Large-scale Model Framework towards Healthcare

Weihua Liu[1], Yong Zuo[1]

[1]AthenaEyes CO.,LTD., Beijing, China

liuweihua@a-eye.cn, zuoyong@a-eye.cn

**Abstract**

In healthcare, multimodal data is prevalent and requires to be comprehensively analyzed before diagnostic decisions, including medical images, clinical reports, etc. However, current large-scale artificial intelligence models predominantly focus on single-modal cognitive abilities and neglect the integration of multiple modalities. Therefore, we propose Stone Needle, a general multimodal large-scale model framework tailored explicitly for healthcare applications. Stone Needle serves as a comprehensive medical multimodal model foundation, integrating various modalities such as text, images, videos, and audio to surpass the limitations of single-modal systems. Through the framework components of intent analysis, medical foundation models, prompt manager, and medical language module, our architecture can perform multi-modal interaction in multiple rounds of dialogue. Our method is a general multimodal large-scale model framework, integrating diverse modalities and allowing us to tailor for specific tasks. The experimental results demonstrate the superior performance of our method compared to single-modal systems. The fusion of different modalities and the ability to process complex medical information in Stone Needle benefits accurate diagnosis, treatment recommendations, and patient care.

## 1. Introduction

In recent years, with the emergence of large language models, including ChatGPT and GPT-4, the field of artificial intelligence has been revolutionized. These models exhibit remarkable fluency in reading, writing, and conversing, even achieving resembling human language proficiency. Concurrently, models, such as residual networks, have also surpassed human performance in some vision tasks, highlighting their exceptional capabilities. However, these single-modal models fall short of achieving the breadth of human intelligence, which can process various modalities of data. For instance, while ChatGPT excels in language-based tasks, it lacks the ability to process or generate visual or auditory content. Conversely, vision-based models like visual transformers or stable diffusion exhibit remarkable visual understanding and generation capabilities, yet they struggle with multi-round task processing. These limitations hinder the large-scale models to be applied in healthcare. Obviously, multiple rounds of multimodal conversations are essential for healthcare, especially in diagnosing a suspected disease.

The previous researches demonstrate the integration of multimodal data offers substantial accuracy enhancements on artificial intelligence models, compared to single-modal approaches for the same tasks. Furthermore, human cognition inherently encompasses multimodality, involving the fusion of natural language, facial movements, acoustic behaviors, and other modalities. Therefore, it is valuable to develop a unified large-scale model framework capable of handling multimodal tasks in practical usage scenarios.

Therefore, in this paper, we introduce Stone Needle, a general multimodal large-scale model framework specifically designed for healthcare. Stone Needle leverages the information of different modalities to enhance the performance in various multi-round dialogue. Our contributions can be summarized as follows:

- By integrating multiple modalities such as text, images, videos, and audio, our method solves the limitations of single-modal systems. This comprehensive approach enables a deeper understanding of medical data while empowering the system to process and generate content from diverse modalities.
- We achieve multi-modal interaction in multiple rounds of dialogue, incorporating both historical dialogues and current inputs and perform intent analysis to select the most suitable medical foundation model. This iterative approach enables effective and interactive conversations, particularly in diagnosing suspected diseases.
- We provide a general framework that ensures the applicability of Stone Needle to a wide range of multimodal medical data. This framework offers scalability and adaptability, allowing healthcare professionals, researchers, and developers to tailor the framework to their specific needs.
- Through experiments, we demonstrate the superior performance of Stone Needle compared to unimodal systems. The fusion of different modalities and the ability to process complex medical information enhance the accuracy of diagnoses, improve treatment recommendations, and elevate the overall quality of patient care.

## 2. Related Work

### 2.1 Multimodal Large Models

Multimodal models aim to integrate multiple modalities, including text, image, audio, and video, to enhance the overall performance of AI systems. Wu et al. [5] introduced Visual ChatGPT, a framework that incorporates visual foundation models to enable user interaction with ChatGPT using images. Huang et al. [10] introduced Kosmos-1, a multimodal large language model for processing multimodal tasks. Liu et al. [6] proposed LLaVA, an end-to-end trained large multimodal model that connects language-only GPT-4 with a vision encoder. Huang et al. [7] developed AudioGPT, which leverages LLMs and foundation models to process complex audio information. These multimodal models leverage the complementary strengths of different modalities, enabling a more comprehensive understanding of input data.

### 2.2 Dialogue Large Language Models

With the advances in Transformer architecture [1], large language models (LLMs)

have garnered significant attention in recent years due to their remarkable capabilities in natural language understanding and generation. Touvron et al. [2] introduced LLaMA, a collection of foundation language models that achieve state-of-the-art performance by training exclusively on publicly available data. Models like GPT-3 [3] and GPT-4 [4] have also demonstrated impressive fluency and comprehension, exhibiting human-like language proficiency in multiple rounds of dialogue. While these models excel in unimodal language tasks, they typically lack the ability to effectively process and generate content from other modalities.

## 2.3 Medical Large Model Frameworks

In the healthcare domain, the development of large-scale models holds the potential to revolutionize medical research, diagnosis, and treatment. Medical large model frameworks focus on integrating medical images, clinical text, and physiological signals to provide comprehensive and accurate insights. Li et al. [8] proposed ChatDoctor, a framework that combined medical datasets finetuned LLMs and image caption models. Wang et al. [9] presented ChatCAD, which integrates LLMs with multiple medical-image computer-aided diagnosis networks by summarizing and reorganizing the information presented in natural language text format. However, the development of comprehensive multimodal frameworks designed explicitly for healthcare remains an ongoing challenge.

In summary, multimodal large models, dialogue large language models, and medical large model frameworks have shown significant advancements in various domains. However, there is still a need for comprehensive multimodal medical large model frameworks that can effectively handle the complexities of healthcare data. The Stone Needle framework proposed in this paper aims to address this gap by providing a general multimodal large-scale model foundation specifically designed for healthcare applications.

## 3. Methodology

Stone Needle is a comprehensive medical multimodal model foundation (MMF) framework as shown in Fig. 1. designed to handle input $q_n$ in various modalities, including text, image, video, and audio. The MMF first leverages historical dialogue $C = \{(q_1, r_1), (q_2, r_2), \ldots, (q_{n-1}, r_{n-1})\}$ and current input $q_n$ to perform intent analysis $\mathcal{T}$. Based on the results of the intent analysis, the framework selects an appropriate medical foundation model (MFM) tailored to specific task, such as medical image analysis (e.g., detection, segmentation, diagnosis), health index prediction, or speech recognition. The processing results and historical dialogue $C$ are structured within the prompt manager $\mathcal{P}$ and subsequently fed into the medical language module (MLM) to generate a response $r_n$. Therefore, the response $r_n$ is formulated as follow:
$$r_n = \text{MMF}(q_n, C)$$
By integrating multimodal inputs and utilizing task-specific MFMs, Stone Needle enables efficient and context-aware processing of medical data, facilitating accurate and personalized responses in medical applications.

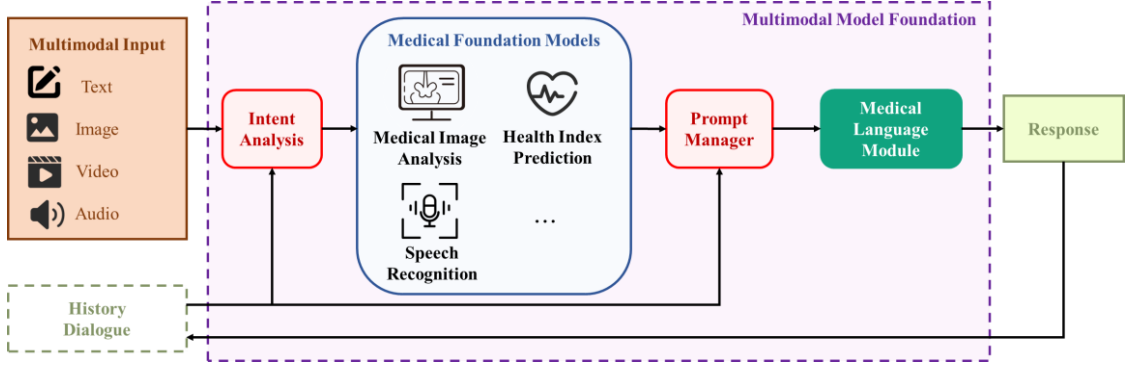

**Fig 1.** Architecture of Stone Needle.

## 3.1 Intent Analysis

In intent analysis phase as shown in Fig. 2, the multimodal input $q_n$ is divided into two distinct parts: text $q_n^{(t)}$ and other resources, $q_n^{(s)}$ (images, videos, audios). Similarly, the query content in the historical dialogue $C$ is also subdivided accordingly. It is important to note that in each query round, either part of the input can be empty ($\emptyset$). The multimodal input at each round $q_k$ is represented as a combination of its textual and other modal resources:

$$q_k = \left\{q_k^{(t)}, q_k^{(s)}\right\}, where\ k \in \{1, 2, \ldots, n\}$$

Next, the intent analysis $\mathcal{T}$ assigns a multimodal base model and determines the suitability based on the textual content and the format of other modal resources in the current input $q_n$ and the queries from historical dialogue $C$.

$$\left(M_i, p_{M_i}\right) = \mathcal{T}(q_n, C)$$

where $M_i$ represents the model of MFMs and $p_{M_i}$ represents the task-related probability associated with that model. Here, $i \in \{0, 1, 2, \ldots, m\}$, where $m$ is the number of MFMs available. $M_0$ signifies the absence of any MFM being selected.

Subsequently, the specific MFM to be used is chosen based on the task-related probability $p_{M_i}$ as follows:

$$\text{MFM} = \max_{p_{M_i}} M_i$$

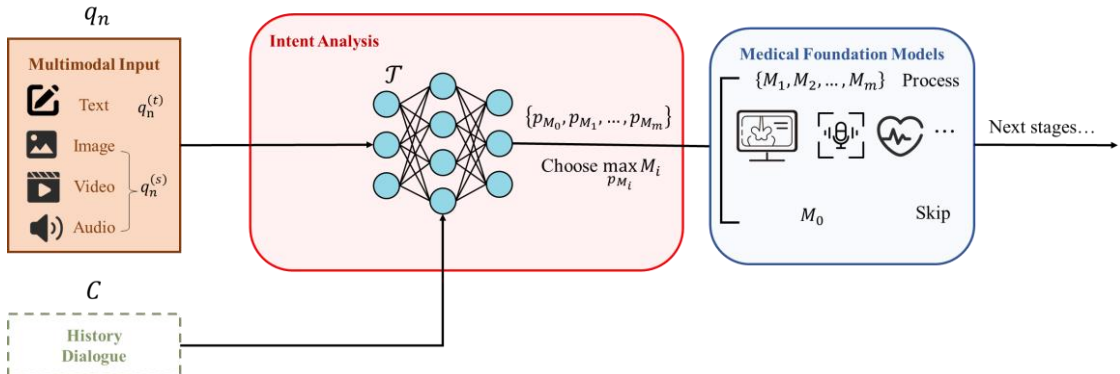

**Fig 2.** Procedure of the intent analysis.

## 3.2 Medical Foundation Models

MFMs refer to model ensembles comprising m distinct task models.
$$\mathrm{MFMs} = \{M_1, M_2, \ldots, M_m\}$$

Once the intent analysis phase has identified the appropriate MFM, the multimodal input is passed to the selected MFM. The output result $R_n$ can take various forms, such as medical image descriptions, health index values, or transcribed voice text.

$$R_n = \begin{cases} \mathrm{MFM}(q_k^{(s)}), & \mathrm{MFM} \neq M_0 \\ \emptyset, & \mathrm{MFM} = M_0 \end{cases}$$

where $k \in \{n, n-1, \ldots, 1\}$. This means that if there are no multimodal resources available for the current query, our framework can automatically retrieve the required resource from nearby queries. Additionally, if no MFM is selected during intent analysis, the framework will proceed to the next phase without further processing.

If the output result $R_n$ is in the form of text, it will be forwarded to the subsequent phase. However, if the output consists of images, audios, or other modalities (e.g., segmented images), they will be returned as resources.

### 3.3 Prompt Manager

The input to the prompt manager $\mathcal{P}$ consists of the output result $R_n$ from the MFM, current input $q_n$ and the historical dialogue $C$. Furthermore, the intent manager in this paper incorporates a structured retriever as an integral part of its functionality. This structured retriever is designed to retrieve relevant attributes associated with the entity from a knowledge base, such as diseases, symptoms, or medical inspections. The retrieved attributes are then utilized to annotate and normalize the corresponding entities within the dialogue. These inputs are combined to form a coherent prompt $P_n$ for further processing.

$$P_n = \mathcal{P}(R_n, q_n, C)$$

By leveraging the knowledge base, the prompt manager enhances the understanding and contextual awareness of the dialogue, as it incorporates structured information. This integration enables improved entity recognition and facilitates the generation of more accurate responses within the context of multimodal medical conversations.

### 3.4 Medical Language Module

The MLM utilized in this framework is composed of the gpt-3.5-turbo, a powerful large-scale language model based on the GPT architecture. The MLM enhances the conversational abilities of the system, allowing it to provide accurate and informative responses within the context of multimodal medical conversations. The input to the MLM is the structured text generated by the prompt manager, which provides the necessary context and information for generating appropriate responses.

$$r_n = \mathrm{MLM}(P_n)$$

To ensure the MLM proficiency in medical understanding and terminology, it undergoes a three-step training process. In the first step, the module is pre-trained on a large-scale dataset of general question-and-answer dialogues. After pre-training, the MLM undergoes fine-tuning using medical data specific to the desired application domain. This fine-tuning process involves training the module on a specialized dataset tailored to the specific medical context, such as electronic health records, medical literature, or other relevant medical sources. This fine-tuning enables the MLM to adapt its knowledge and responses to the specific nuances and requirements of the medical domain. Furthermore, the MLM also undergoes fine-tuning through medical question-and-answer dialogues. This training involves exposing the module to a dataset comprising real-world medical queries and their corresponding expert answers. By fine-tuning on this dataset, the MLM gains the ability to generate responses that are aligned with expert medical knowledge and best practices. The pre-training and fine-tuning processes enable the MLM to leverage its comprehensive understanding of medical language and knowledge, allowing it to provide accurate, context-aware, and reliable responses within the Stone Needle framework.

## 4. Experiment

### 4.1 Experimental Setup

We conducted an experimental study using the Stone Needle framework in a healthcare setting. We utilized a multimodal dataset comprising anonymized patient records. It contains 865,275 text, 428,391 images, 421,785 videos and 463,814 audio for training while 1,000 text, 1,000 images, 1,000 videos and 1,000 audio for evaluation. The experiments are performed on a Tesla A100 GPU for efficient computation. The prompt manager components of the framework are trained on the dataset to enable the recognition and retrieval of pertinent attributes related to medical conditions, symptoms, and treatments. This training process enhanced the prompt manager's ability to understand and respond to queries effectively. Additionally, the MLM was pre-trained on a large-scale medical question-and-answer dialogue dataset, improving its understanding of medical terminology and facilitating more accurate responses.

To evaluate the performance of the Stone Needle framework, we compared its diagnoses and treatment recommendations with those provided by healthcare professionals. Accuracy served as the primary measure for assessing the effectiveness of the framework. We employed a rigorous evaluation process to ensure a fair and unbiased comparison between the Stone Needle framework and human experts. Throughout the evaluation, we recorded and analyzed the accuracy of the framework's diagnoses and treatment recommendations, as well as the corresponding outcomes provided by healthcare professionals. In order to quantify the accuracy and effectiveness of the Stone Needle framework, we compared its results with existing open-source large models.

### 4.2 Quantitative Evaluation

The quantitative evaluation of Stone Needle was conducted to assess its performance and effectiveness in handling multimodal medical data and providing accurate diagnoses and treatment recommendations. Several key metrics were used to evaluate the system, including accuracy, precision, recall, and F1 score. To measure the accuracy of Stone Needle, a dataset of real-world medical cases was used. Each case consisted of multimodal input data, including text, images, videos, and audio, along with corresponding diagnoses and treatment recommendations provided by healthcare professionals. Stone Needle's responses were compared against the ground truth to determine the accuracy of its diagnoses and treatment suggestions. We employ a cross-validation approach, randomly dividing the dataset into training and testing sets.

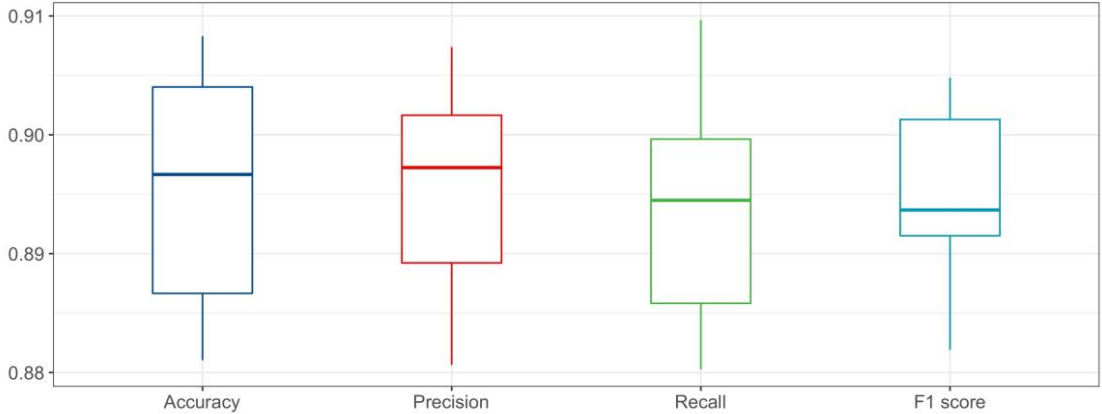

**Fig. 3.** Quantitative evaluation of Stone Needle

The quantitative evaluation of Stone Needle provides objective measures of its performance in handling multimodal medical data and delivering accurate diagnoses and treatment recommendations, as shown in Fig. 3. These metrics enable a comprehensive assessment of the system's effectiveness and contribute to advancing the field of AI in healthcare.

### 4.3 Comparison

We compare Stone Needle with several existing models, including GPT-4, LLaMA, Visual ChatGPT, and LLaVA to evaluate its performance and superiority in the medical domain, as shown in Table 1. Compared to GPT-4 and LLaMA, our method has higher accuracy in unimodal language tasks with capabilities in handling multimodal medical data. This multimodal integration allows Stone Needle to surpass the capabilities of unimodal large language models in the medical domain. On the other hand, LLaVA and Visual ChatGPT are on incorporating visual information into conversational AI systems. It enables users to interact with the model using images and text. However, Stone Needle enhances its capabilities in handling multimodal medical data including text, images, videos, and audio, allowing it to deliver more comprehensive and accurate responses.

**Table 1.** Comparison with other methods.

| Model | Text | Image | Video | Audio |
|---|---|---|---|---|
| GPT-4 [4] | 0.8487 | / | / | / |
| LLaMA-7B [2] | 0.7638 | / | / | / |
| Visual ChatGPT [5] | 0.8331 | 0.6273 | / | / |
| LLaVA [6] | 0.8472 | 0.6177 | / | / |
| Stone Needle(ours) | 0.9063 | 0.8974 | 0.8942 | 0.8955 |

The experimental results clearly demonstrate the superiority of Stone Needle in the medical domain compared to GPT-4, LLaMA, Visual ChatGPT, and LLaVA. By effectively integrating multiple modalities and specifically addressing the needs of healthcare applications, Stone Needle can provide healthcare professionals with valuable insights and improving patient care.

## 4.4 Case Studies

The case studies provide valuable insights into the performance of the Stone Needle framework in a real-world healthcare environment. These studies showcase the diverse capabilities of Stone Needle in addressing a range of medical scenarios by leveraging multiple modalities, demonstrating its potential to enhance healthcare practices and empower users to manage their health effectively. The details are presented in the appendix.

# 5. Conclusion

In conclusion, this paper presented Stone Needle, a general multimodal large-scale model framework specifically designed for healthcare applications. By addressing the limitations of current AI models that predominantly focus on single-domain cognitive abilities, Stone Needle bridges the gap between unimodal and multimodal AI/ML systems in the healthcare domain. Through its components of task analysis, medical foundation models (MFMs), prompt manager, and medical language module (MLM), Stone Needle enables the integration of various modalities such as text, image, video, and audio. This multimodal approach enhances the capabilities of AI/ML systems, empowering healthcare professionals to effectively tackle complex real-world challenges. The integration of multimodal data in healthcare has the potential to revolutionize diagnosis, treatment, and patient care. By effectively leveraging diverse modalities, it contributes to the development of more comprehensive, accurate, and human-like healthcare solutions. The framework holds immense potential for transforming the way healthcare is delivered and has far-reaching implications for improving patient care.